\documentclass{bmvc2k}

\usepackage{bm}             
\usepackage{booktabs}       
\usepackage{amsmath}        
\usepackage{amsfonts}       
\usepackage{nicefrac}       
\usepackage{microtype}      
\usepackage{adjustbox} 
\usepackage{booktabs}

\title{Human-like Relational Models for Action Recognition in Video}

\addauthor{Joseph Chrol-Cannon}{jc0080@surrey.ac.uk}{1}
\addauthor{ Andrew Gilbert}{https://www.surrey.ac.uk/people/andrew-gilbert}{1}
\addauthor{Ranko Lazic}{R.lazic@surrey.ac.uk}{1}
\addauthor{Adithya Madhusoodanan}{adithya.madhusoodanan3@gmail.com}{2}
\addauthor{Frank Guerin}{https://www.surrey.ac.uk/people/frank-guerin}{1}

\addinstitution{
 University of Surrey\\
 Guildford, UK
}
\addinstitution{
 National Institute of Technology Karnataka\\
 Mangalore City, \\
 Karnataka State, India,
}

\runninghead{Student, Prof, Collaborator}{BMVC Author Guidelines}

\begin{document}

\maketitle

\begin{abstract}
Video activity recognition by deep neural networks is impressive for many classes. However, it falls short of human performance, especially for challenging to discriminate activities. Humans differentiate these complex activities by recognising critical spatio-temporal relations among explicitly recognised objects and parts, for example, an object entering the aperture of a container. Deep neural networks can struggle to learn such critical relationships effectively. Therefore we propose a more human-like approach to activity recognition, which interprets a video in sequential temporal phases and extracts specific relationships among objects and hands in those phases. Random forest classifiers are learnt from these extracted relationships. We apply the method to a challenging subset of the something-something dataset~\cite{goyal2017something} and achieve a more robust performance against neural network baselines on challenging activities.
\end{abstract}

\begin{figure}[htp]
\centering
\includegraphics[width=\textwidth]{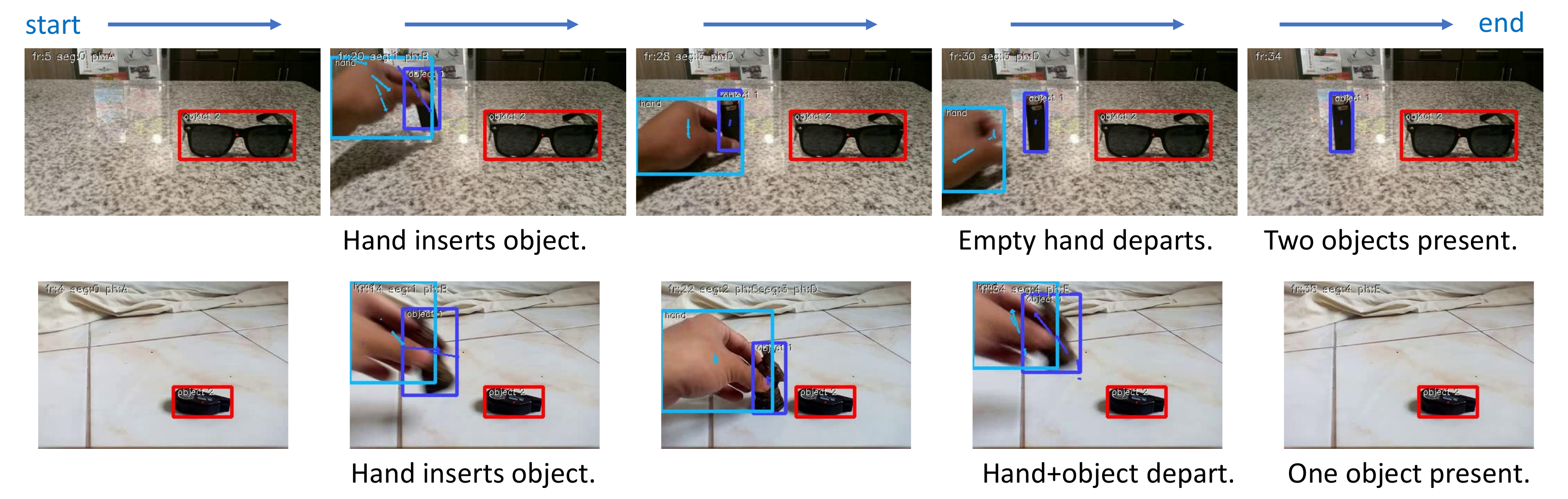}
\caption{In the something-something dataset~\cite{goyal2017something} some actions have very similar visual and motion features: Top: action ``putting something next to''; Bottom: action ``pretending to put something next to''. These actions are easily confused by deep learning approaches (See confusion matrix Fig.~\ref{fig:confusion}). We can overcome this if we can identify the temporal segments where key discriminating events happen and then look at the bounding boxes of objects and hand and their motion and relations (e.g. contact) in these segments. These critical features can discriminate these easily confused categories. }
\label{fig:teaser}
\end{figure}

\section{Introduction}
Current approaches to video activity recognition are generally based on deep learning networks that internally learn local and global features. However, superficially similar activities will often confuse these aproaches. For example, in the AVA-Kinetics Challenge 2020, the winning mean average precision was 39.6\%, and for the most challenging classes, average precision ranged from 3\% to 14\% \footnote{https://research.google.com/ava/challenge.html}. Discriminating between such activities is of great importance in many applications of significance to society, for example, monitoring older adults in their residences and alerting them if an event requiring intervention has occurred. Or to discriminate taking medicine from drinking water, or differentiate pouring in a container from spilling, possibly causing an electric or falling hazard. Also, in urban surveillance, to discriminate an assault from a friendly hug or a theft from a legitimate object transfer.

Research has shown \cite{BenYosef2018ImageIA} that humans discriminate similar activities by differences in relationships among critical parts of objects, such as the position of the hand relative to another person's back, differentiating between fighting and hugging. While such relationships could, in principle, be learnt by a deep learning approach, in practice, deep learning tends not to make use of features that are easily interpretable by humans, as the training process would need to be conditioned to take human characteristics into account. 

Therefore in this work, we take an opposite approach, a more top-down based approach, using only a minimal number of higher level elements extracted from the video that are more meaningful to humans. Specifically, we use only the bounding boxes of critical objects and hands in the video frames, without any other visual features or optical flow. From this, we compute relationships among these essential objects and learn these patterns via machine learning. Our goal is to investigate the performance that we could achieve by only using this high-level interpretable information. Ultimately we can fuse this complementary information with deep learning features for the most robust performance. 

\section{Related Work}

Current leading approaches to video activity recognition include Temporal Shift Module \citep{9008827}, SpatioTemporal and Motion Encoding  \citep{9010925}, R(2+1)D \cite{8578773}, D3D \citep{Stroud_2020_WACV}, TRN \cite{Zhou_2018_ECCV}, and Temporal  Reasoning Graphs \citep{9062552}. These techniques use bottom-up features derived from networks for object recognition, short-range temporal information, and motion features, but no high level models explicitly capturing relationships among parts. Human visual processing, in contrast, seems to use models that explicitly capture relationships among significant parts \cite{BENYOSEF201865,BenYosef2018ImageIA}. We also see significant differences in the accuracy with which humans recognise activities, especially for low quality or similar action videos \cite{BENYOSEF2020104263}.
For this reason, there is increasing interest in augmenting the deep learning architectures with the kinds of features that humans seem to use. For example, \citet{Wu_2019_CVPR} add long term features from regions of interest. Recent work by \citet{CVPR2020_SomethingElse} augments pure bottom-up deep learning with the explicit inclusion of detected bounding boxes for the critical objects in the scene and object features for those objects. These are concatenated with standard video features from pre-existing architectures such as I3D and STRG \cite{wang2018videos}. However, how these meaningful bounding boxes are used is determined by the black box neural network; we do not have an interpretable explanation of why the video is determined to be in a particular class. Such interpretability can be essential to debugging erroneous classifications and making the system learn more human-like models with higher accuracy.

For still images \citet{BENYOSEF201865} presented a human-like approach that learns models for specific classes; the models can then be applied top-down to produce an interpretation of that image. The model contains parts and relationships among them; an interpretation maps model parts to regions of the image. Unfortunately, this same approach has not been applied to video to the best of our knowledge. In a video, it would be necessary to map significant parts in frames and map temporal segments of the video to temporal phases of the activity.

\section{Methodology}

\subsection{Methodology Overview}
\begin{figure}[htp]
\centering
\includegraphics[width=\textwidth]{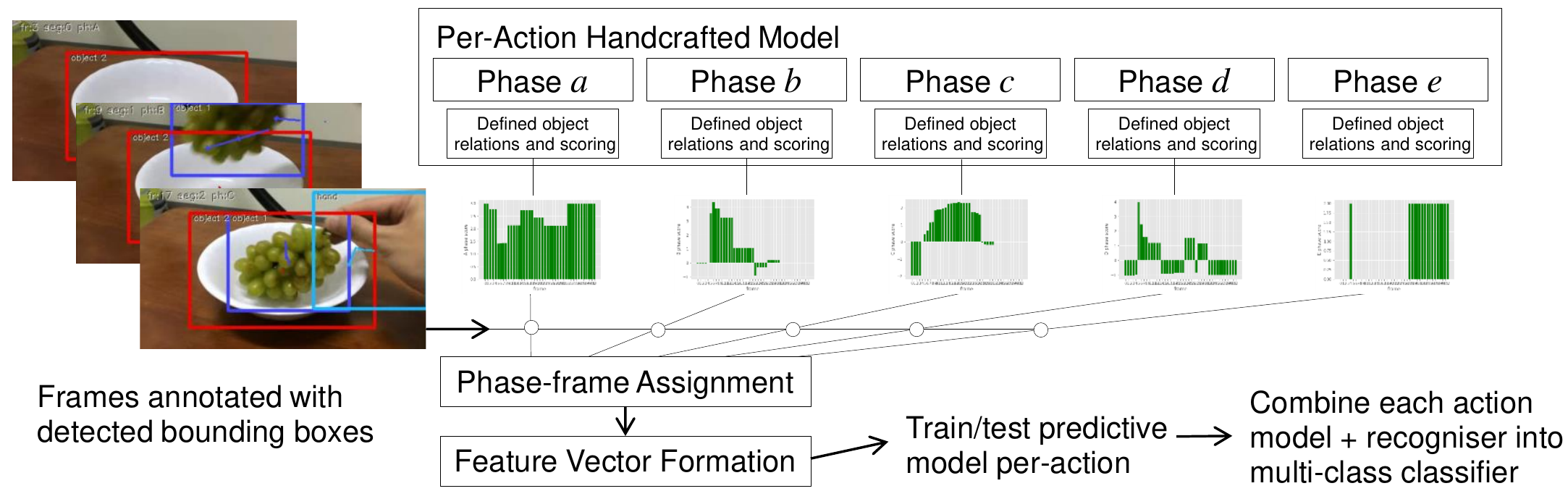}
\caption{Overview of the proposed top-down recognition model. Sequential frames are
processed by a set of hand-crafted action models that consist of object relations and scoring
for each phase of an action. Feature vectors are formed from the object relations of phase-assigned
frames. These are used to train a set of random forests}
\label{fig:architecture}
\end{figure}
The high-level overview of our system\footnote{we will provide code repository link in the final paper version.} is depicted in Fig.~\ref{fig:architecture}. For our top-down interpretable mode, we use as an input only the bounding boxes of the principal objects and hands from the video frames (there are no lower level visual or optical flow features from the videos). We interpret the video as a possible member of each of the 17 action categories using 17 action specific models (described below). We temporally segment the video into five phases for each action category, representing the sequential stages involved in that action. We then compute feature vectors characterising each phase and train a random forest classifier for each category (with the positive examples being that category and other categories being negative examples). When doing multi-class classification, the highest probability random forest prediction is returned as the class.

\subsection{Assigning Action Phases and Computing Phase Features}

\begin{figure}[htbp]
\centering
\includegraphics[width=\textwidth]{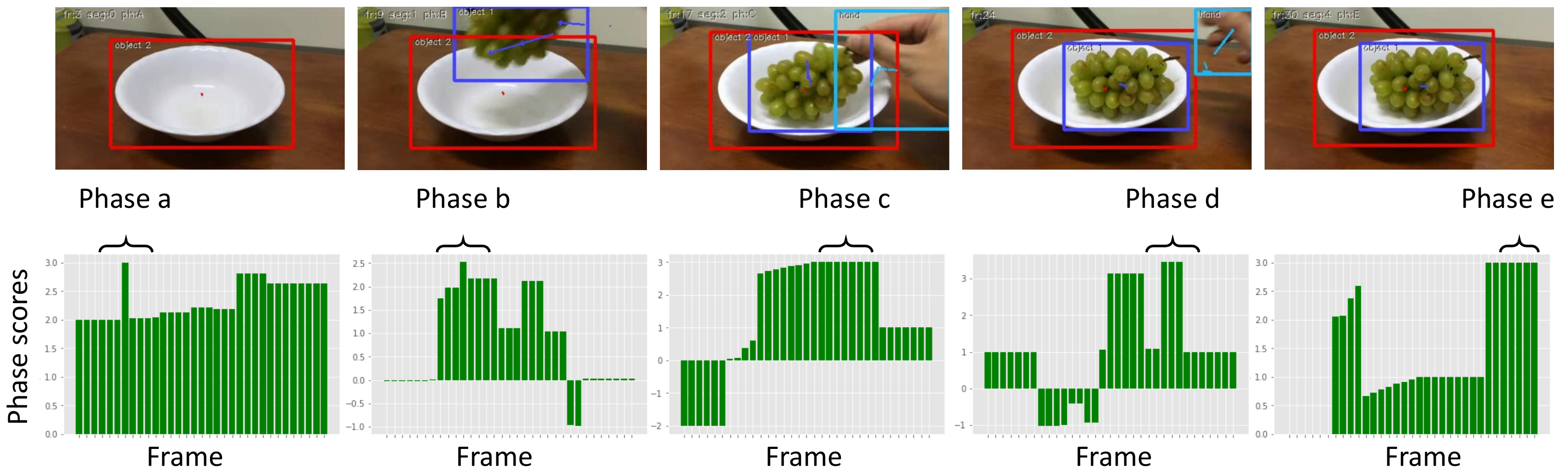}
\caption{Top: Activity frames illustrating the 5 phases for the action 'Putting something into something'. (red and blue bounding boxes indicate objects and light blue the hand) \\
Bottom: Phase scores for all frames in an example video. For example, the second bar chart depicts applying phase $b$ scoring across all frames; we can see that the beginning and end regions give low scores for phase $b$, while the portion indicated by the top bracket gives the maximal response. Note: due to different
scales between phases, any comparisons in the phase assignment algorithm are made within the same phase.}
\label{fig:phase_score}
\end{figure}

In a traditional deep learning approach, a model architecture would be handcrafted and trained. Instead, for each activity, we propose to handcraft a `model' which interprets the video. 
We identified that all actions could be segmented (temporally) into five broad phases, with a characteristic difference in each phase. The phases are labelled $a,b,c,d$ or $e$ and a typical segmentation is illustrated in Fig.~\ref{fig:phase_score} (top).
\vspace{-1ex}
\begin{itemize}
\item Phase $a$: The object(s) is present in the scene, the manipulation has not happened;
\vspace{-1ex}
\item Phase $b$: The hand enters (possibly carrying an object);
\vspace{-1ex}
\item Phase $c$: The critical manipulation happens (e.g. object placed or picked);
\vspace{-1ex}
\item Phase $d$: The hand departs (possibly carrying an object);
\vspace{-1ex}
\item Phase $e$: The objects are present, with the result of the manipulation evident.
\end{itemize}
\vspace{-1ex}
Most videos depict all five phases, but some do not; e.g. some videos start at phase $c$, with the hand already contacting the moved object, while other videos finish at $c$ without sufficient frames to make a clear $d$.

An activity model consists of a scoring function for each of the five phases.  A scoring function for a phase $p$, for activity $\mathit{act}$, maps each frame of a video to a real-valued score which is a heuristic to rate how well that frame matches the typical features expected of a frame in phase $p$ of $\mathit{act}$.
Features used for scoring (and also eventually used to characterise the phase) include binary values such as the presence of objects and hand, and whether they are moving, and real-valued features such as the relative motion between objects or object and hand (see Table~\ref{tab:relations}). 

\begin{table}[hbp]
\centering

\begin{tabular}{ l }
 \toprule
 
 Size($O_1$) = width $\times$ height  \\
 
OverlapNormalised($O_1$,$O_2$) = (x overlap $\times$ y overlap) $/$ (0.1$\times$size of smaller box) \\
Offset($O_1$) = centre in current frame - centre in previous frame \\ 
OffsetDist($O_1$,$O_2$) = difference in offsets of $O_1$, $O_2$ \\
OffsetAngle($O_1$,$O_2$) = difference in offset angles \\ 
CentreDist($O_1$,$O_2$) = difference in centres  \\ \midrule
\textbf{Binary relations}:
Present($O_1$), 
 Touching($O_1$,$O_2$),  
 Contained($O_1$,$O_2$), \\
 CentreOnTop($O_1$,$O_2$),  
 CentreUnderneath($O_1$,$O_2$), \\
MoveWithHand(..), 
HandMoveRelative(..),  
ObjectMoveRelative(..) \\
 \bottomrule
\end{tabular}
\caption{The functions used to compute relational features between bounding boxes of objects and
hands within each frame of video. $O_1$ and $O_2$ are two objects annotated with bounding boxes. As in  Fig.~\ref{fig:phase_score} (top), $O_1$ is the stationary object while $O_2$ is the moving one, but the system tries each video with the order swapped as well, picking the best match. `Offset' and `Move' features are computed across two consecutive frames. The Move features are boolean, true if movement exceeds a threshold.}
\label{tab:relations}
\end{table}

Given a video we want to interpret, the scoring functions compute a `phase score' for each video frame. Thus, the scoring function describes how well each frame matches the expected features for each phase. 
The phase scores across frames of a sample video are depicted in Fig.~\ref{fig:phase_score} (bottom).
The score for phase $b$ peaks around a quarter of the way through the sequences when the grapes move into the bowl. While phase $c$ peaks during the stationary portion when the hand holds the grapes in the bowl before withdrawing in $d$, phase $e$ is maximal in the later frames when the action is complete, and the result is evident.

\begin{figure}[!h]
\includegraphics[width=6.3cm]{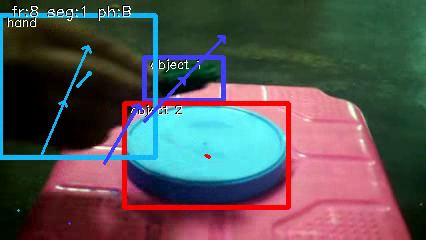}
\caption{In this example  Object 1's trajectory is away from the centre of Object 2, during phase $b$. In general a short segment of video cannot unambiguously be assigned a phase; instead we need to find a best global assignment for the video, meaning that the segment pictured is effectively having phase $b$ imposed on it top-down, because that makes a good global assignment.   }

\label{moveaway}
\end{figure}

Given the frame-wise phase scores, a Gaussian kernel is convolved with the sequence of scores for each phase to identify the maximal response. This procedure helps to overcome noise in the phase scores. For example, the bounding boxes can be noisy when the annotation (by a crowd-worker) is copied from a previous frame but eventually is adjusted to the new appropriate size, causing an extreme and noisy increase in the vector of the box's motion. Also, sometimes the hand may briefly take an object away from the centre of the central object during phase $b$, especially if the main object is large and the hand needs to move up to put an item in or on, see Fig.~\ref{moveaway}. 

There is a strict order  $a,b,c,d,e$  to phases in an activity. Assigning frames to phases is not simple; e.g. if we were to first assign phase $a$ in   Fig.~\ref{fig:phase_score} we would assign it near the end of the video because the gaussian has the highest response there; this is a wrong assignment. In general, the scores for any particular phase could be misleading, but most of the phase scores are likely accurate. Therefore we want to do an assignment that respects the strict order while maximising the total matched phase scores across the whole video.
We approached this in a simple way that is not guaranteed to be optimal.

We identify the centre frame $f$ for each phase based on the highest score responses. However we perform this in a greedy fashion in the phase order: $b,a,d,c,e$ i.e. $f_b$ the centre frame of the $b$ phase is assigned first, then $f_a$ is assigned to the highest  response region in frames preceding $f_b$, $f_d$ is the highest response in frames succeeding $f_b$, etc. This order is used as we heuristically found that phase $b$ is a key anchor to the sequence. 
Each phase consists of the central frame and three frames on each side, unless that would overlap another phase, in which case the phase stops at the midpoint.

We also did an alternative assignment picking the second best response for $b$, because sometimes the highest response $b$ is not the correct $b$, and if so, assigning that to $b$  and restricting $c, d, e$ to subsequent frames will produce a poor assignment (with low total score). To find the second best $b$, we remove the best $b$ from the sequence and three frames on each side.
The complete assignment using the best $b$ and the complete assignment using the second best $b$ are scored to find a winner. It could be the third or fourth best $b$ is the correct one, But we found going as far as the second was adequate.

Finally, this entire procedure was repeated for swapped assignments of object 1 and object 2 (because we cannot assume that the annotated object 1 is the primary object doing the movement), creating four alternative assignments, each of which is scored the winner chosen.

\subsection{Phase feature vector}

To construct a feature vector to represent each phase. The detected central frames, and a window of +/- $n$ frames (generally $n=3$) are selected to represent each phase $p$. The phase feature vector $F_p$ is defined as the concatenation of four data statistics over the phase scores $s$. 

\begin{equation} \label{PhaseVectorEqn}
	F_p=\{s_{mean},s_{med},s_{max},s_{min}\}
\end{equation}
Where $s_{mean}$ is the mean of the $2n+1$ frame scores, $s_{med}$ is the median and $s_{max}$ and $s_{min}$ and the maximum and minimum scores over the selected frames. This is repeated for all the phases $a,b,c,d,e$ to create a feature vector embedding $V$ for a video sequence of $V=\{F_a,F_b,F_c,F_d,F_e\}$ which is the input to a random forest classifier.


\subsection{Activity Recognition with Random Forests}
To classify the video sequences correctly, given the feature embedding created by the phase ordered object relations $V$, we trained a random forest classifier~\cite{Breiman}. We also tried Gradient Boosting methods and Neural Networks (MLP), but Random Forests performed best. 
We identified that due to the responses in $V$ varying dramatically in scale and sparsity that even scaling each feature by its standard deviation, MLP based neural networks were out-performed by an ensemble-based random decision forest (RDF). An RDF is an ensemble of random decision trees (RDTs), which considerably improves the stability and accuracy of individual RDT. Ideally suited to diverse features as the decision tree nodes define splits on each feature without combining them.

A Decision Tree is a tree (and a type of directed, acyclic graph) in which the nodes represent decisions or terminal nodes. The edges or branches are binary (yes/no, true/false), representing possible paths from one node to another. The root or topmost node of the tree (and there is only one root) is the decision node that splits the dataset using a variable or feature that results in the best splitting metric evaluated for each subset or class in the dataset that results from the split. According to the splitting metric at each decision node, the decision tree learns by recursively splitting the dataset from the root onwards (in a greedy, node by node manner). The terminal nodes are reached when the splitting metric is at a global extremum.

The Feature vectors are defined per action and so vary in length between different activities. Because of this, rather than have one multi-class model, we fit a binary-class random forest on each activity that detects its presence. Then at test time, each video is run on all binary classifiers, each producing a probability of the action. Thus, the multi-class problem is resolved simply by taking random forest with the highest probability on the video.

\section{Results}
The 20bn project "something-something" data set is an extensive, crowd-sourced collection of over 220K videos, each depicting one of 174 actions being done on one or more objects. The data set is introduced in \cite{goyal2017something} which also provides a CNN model to recognise the actions that we take as a \textbf{baseline CNN}. The dataset is challenging as the action classes can be very similar and provide extensive confusion to algorithms. The "something-else" project \citep{CVPR2020_SomethingElse} extends upon the baseline by introducing annotations to the data set in the form of defined and detected object and hand bounding boxes. As our model involves human-defined abstract features, we directly make use of these annotations. There are up to five objects,  which we call Object 1 ($O_1$), Object 2 ($O_2$), etc., but for the action categories we analysed, only two were needed.  We also run comparisons with the "something-else" model that combines CNN features with the bounding boxes in a more general way (we refer to as \textbf{CNN+bbox}).

We focus on a subset of the dataset, focusing on the set of related 'Putting' and 'Taking' actions such as \emph{Pretending to put something next to something} and \emph{Putting something in front of something}. These are particularly hard to classify for the existing works, and therefore we wish to focus our novel approach around this area. In total there are a 17 actions(class ids: 14, 72, 73, 74, 76, 77, 83,  105, 106, 107, 108, 111, 113, 114, 119, 122 and 149) the description of the actions are listed in the supplementary material. This subset contains over 16K examples sequences (~800,000 frames) for training and 1.6k for validation. so as to compare like-for-like we re-train baseline (VGG-style 3D CNN with 11 layers)~\cite{goyal2017something} and CNN+bbox~\cite{CVPR2020_SomethingElse} with this subset. 

The object bounding box annotations are taken from~\cite{CVPR2020_SomethingElse} and use the weighted average precision as the metric for success as in common when reporting results for this dataset. In Tab~\ref{tab:precision} the average precision across the \emph{something-something} subset dataset for the specific actions are shown together with the retrained \textbf{baseline CNN} and Temporal Relational Networks (\textbf{TRM)} - however TRM was trained on the complete dataset. We also show the performance of the fusion of our work with the baseline CNN approach of~\cite{goyal2017something}, \textbf{ Fused:Proposed+~\cite{goyal2017something}}, and fusion with the work of CNN+bbox~\cite{CVPR2020_SomethingElse}, \textbf{Fused:Proposed+~\cite{CVPR2020_SomethingElse}}.

\begin{table}[hbp]
\centering
\begin{adjustbox}{width=\textwidth}
\begin{tabular}{ l|rrrrrrrrrrrrrrrrrr }
 \toprule
 Smth Smth Class                       & 14 &72  &73  &74  &76  &77  &83  &105 &106 &107 &108 &111 &113 &114 &119 &122 &149  &mAP  \\ \midrule
 Baseline CNN~\cite{goyal2017something}&0.25&0.06&0.20&0.31&0.11&0.17&0.12&0.31&0.72&0.44&0.47&0.44&0.25&0.31&0.28&0.60&0.61 & 0.42\\
 TRM~\cite{Zhou_2018_ECCV}             &0.21&0.06&0.30&0.45&0.23&0.10&0.11&0.54&0.67&0.40&0.61&0.55&0.27&0.33&0.28&0.53&0.57 & 0.37\\
 Proposed                              &0.11&0.07&0.30&0.27&0.20&0.02&0.41&0.32&0.70&0.56&0.64&0.25&0.38&0.32&0.27&0.69&0.74 & \textbf{0.50}\\ \midrule
 Fused:CNN+bbox~\cite{CVPR2020_SomethingElse}&\textbf{0.39}&0.02&0.23&0.24&0.31&\textbf{0.20}&0.45&\textbf{0.48}&0.68&0.55&0.60&\textbf{0.62}&\textbf{0.47}&0.22&\textbf{0.36}&0.69&0.74& 0.42\\ \midrule
 Fused:Proposed+\cite{goyal2017something}    &0.30&\textbf{0.15}&0.38&\textbf{0.43}&0.33&0.11&0.46&0.40&\textbf{0.77}&0.49&0.61&0.43&0.37&\textbf{0.39}&0.34&\textbf{0.72}&0.70 & 0.52\\
 Fused:Proposed+\cite{CVPR2020_SomethingElse}
 &\textbf{0.39}&0.06&\textbf{0.42}&0.39&\textbf{0.39}&0.11&\textbf{0.56}&0.47&0.68&\textbf{0.56}&\textbf{0.70}&0.45&0.46&\textbf{0.39}&\textbf{0.36}&0.67&\textbf{0.78} & \bf{0.57}\\
 
 \bottomrule
\end{tabular}
 \end{adjustbox}
\caption{Precision rates for each method tested on the validation set the Something Something 'hard' subset of video sequences.}
\label{tab:precision}
\end{table}

\begin{figure}[htp]
\centering
\includegraphics[width=\textwidth]{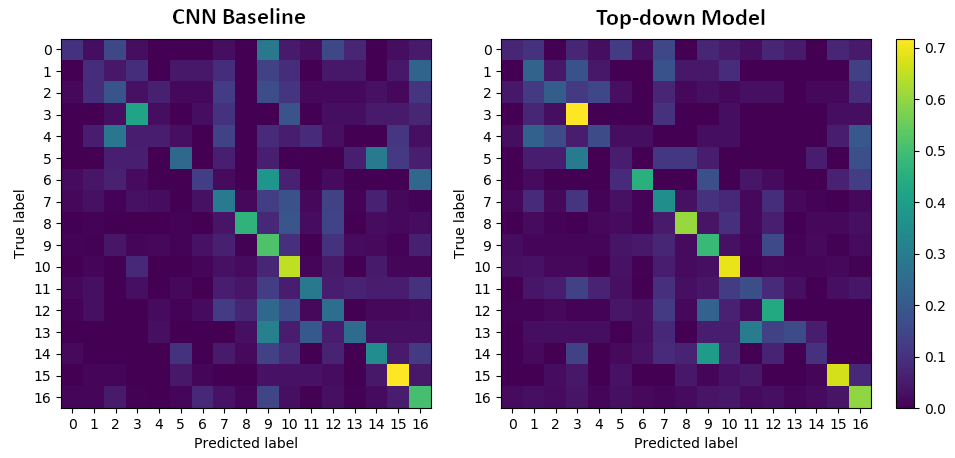}
\caption{Confusion matrices comparing our top-down model against the CNN baseline. The numbered classes 0-16 match in sequence the something-something classes in Table~\ref{tab:precision}}
\label{fig:confusion}
\end{figure}

Our proposed relational model's approach  outperforms the Baseline CNN,  but there are significant differences in the samples they classify correctly, as shown in Fig.~\ref{fig:confusion}. 
This performance difference suggested the possibility of a combined, fused classifier of our proposed relational classifier and the more common CNN based low-level feature work of ~\cite{goyal2017something}. We combine our model with the predictions from the baseline CNN, the pre-threshold probability of the baseline model is summed with ours before selecting the overall highest. The Combined approach improves on the current state the art~\cite{CVPR2020_SomethingElse}, by 2\% across the 1600 videos. Similarly, when comparing the performance of the fused model and the current sota~\cite{CVPR2020_SomethingElse}, there are significant differences across categories, for example, for actions 14, 72 and 74. This demonstrates a substantial difference in the type of relationship that the approach can capture with the human-like approach. 
\begin{figure}[htp]
\centering
\includegraphics[width=\textwidth]{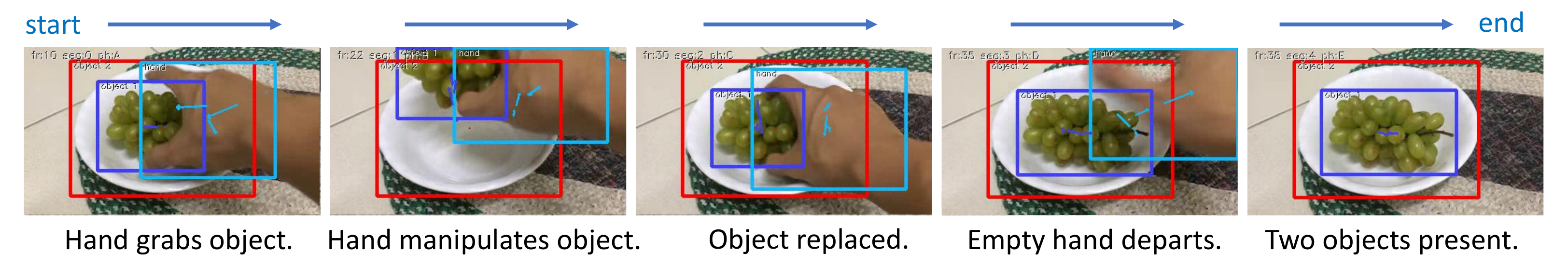}
\caption{Phase assignment to ``pretending to take something out of something''.}
\label{fig:qualitaiveexamples}
\end{figure}

\begin{figure}[htp]
\centering
\includegraphics[width=10cm]{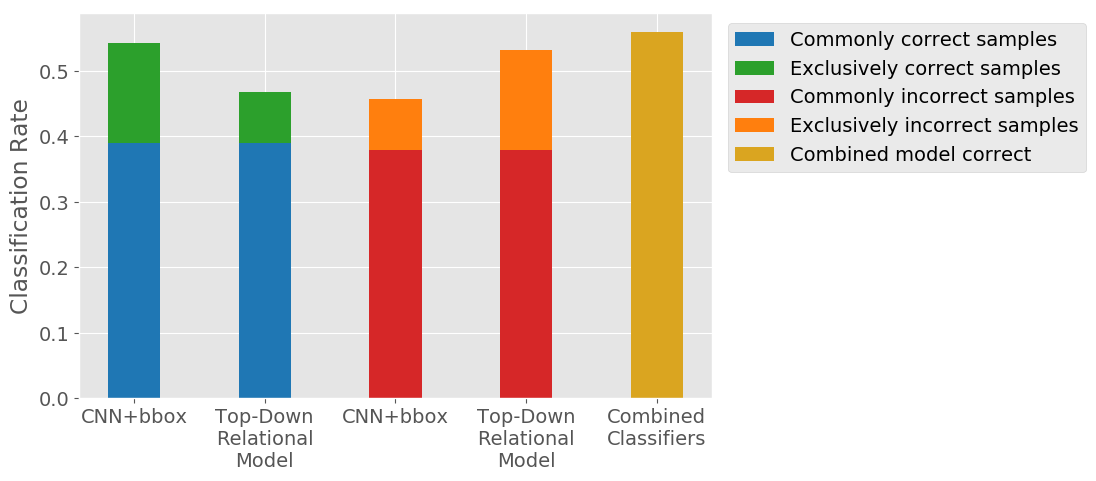}
\caption{Classification rate (overall, not mAP) for our top-down model versus the CNN baseline. Commonly and exclusively correct samples show the extent of diversity between the classifiers.}
\label{fig:agreement}
\end{figure}

\begin{figure}[htp]
\centering
\includegraphics[width=\textwidth]{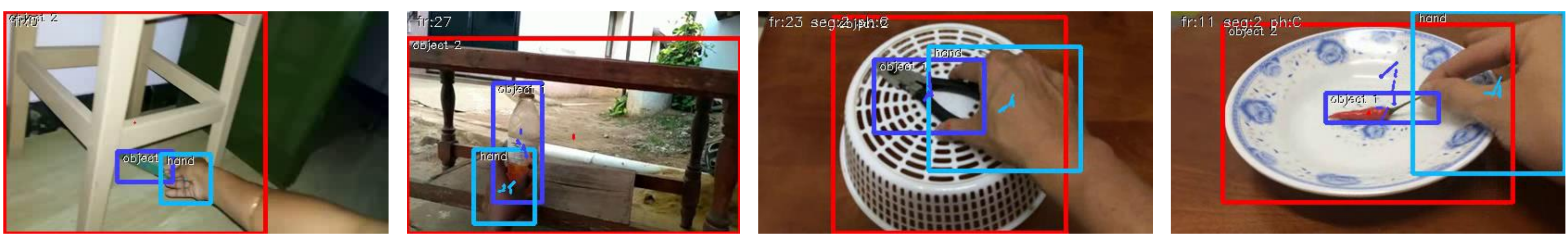}
\caption{Selected images of highly confused categories, see text for details.}
\label{fig:errors}
\end{figure}

Comparing the confusion matrices for our approach and the CNN+bbox~\cite{CVPR2020_SomethingElse}, we see that the latter suffers significant confusion between ``putting something next to'' and ``pretending to put something next to''. At the same time, ours is most decisive in these categories. The phase assignment of our approach on samples from these categories is shown in Fig.~\ref{fig:teaser}. Even though CNN+bbox uses the same bounding boxes as ours, how these are used within its deep learning network somehow fails to capture the critical relationship.
A further example where ours is less confused than                CNN+bbox is                                      ``pretending to take something out of something'', depicted in Fig.~\ref{fig:qualitaiveexamples}. This is a difficult category because of the variety of ways in which crowd workers have chosen to interpret `pretending' for this action. Some take the object and return it, some merely pass over it, and some make several passes. However, there is at least one simple and clear commonality: the object is always left at the end. Our approach with sparse features makes it easier for the learner to focus on such things.

\subsection{Error Analysis}

The interpretable approach makes it easier for us to diagnose errors. Our worst performing category is ``Pretending to put something underneath something''. Inspection of the phase assignments shows difficulties with assigning phase $b$ due to the prevalence of substantial $O_2$ objects in this category (See red bounding box in Fig~\ref{fig:errors}, left; i.e. the object that something is put under); often all of the action happens when the hand and $O_1$ are wholly contained in $O_2$. The heuristics that worked well to identify $b$ in other categories included bounding boxes coming into contact and overlapping boxes increasing; these fail here. The same issue causes significant confusion in Fig.~\ref{fig:confusion} where ``Putting something underneath something'' (see Fig~\ref{fig:errors} second from left) is being mistaken for ``Putting something into something''.

A significant limitation of our approach is a lack of 3D information. For example, to discriminate `into' from `onto' would be difficult for humans if looking only at bounding boxes, especially when the camera views from a high angle (see Fig~\ref{fig:errors}, two images to the right). The only discriminating feature is that the stationary object $O_2$ has higher walls (is a container). Sometimes objects are moving closer together in 3D space. Still, in 2D bounding boxes, they move away because the video starts with one object held near the camera, and as it moves away from the camera, its bounding box gets much smaller, and the centre of the box moves away from the target (in 2D).  Sometimes as an object $O_1$ is withdrawn, it is taken across the front of $O_2$, so in 2D it looks like it is getting closer during the withdrawal phase $d$. Furthermore, videos in some classes  can only be discriminated  by understanding that the camera looks down or from the side, e.g. for `putting something underneath' versus `putting something behind'. A human clearly infers 3D from the 2D videos, and uses this to interpret what is happening. Inferring  information such as whether surfaces are vertical or horizontal in 3D space would be very challenging in the types of close up videos common in this dataset.

\section{Conclusion}
We have presented an alternative novel method to perform activity recognition on challenging similar videos. We took inspiration from a more human-like approach utilising object bounding boxes and created handcrafted rules to identify the phases of the activity. These features extracted from the identified phases  are then trained through a standard random forest classifier to provide state of the art performance on the challenging subset of the Something Something dataset. In future, we aim to improve this performance further through incorporating 3D depth detail and adding additional high level human inspired features such as object recognition.

\bibliography{egbib,refs}

\begin{thebibliography}{14}
\providecommand{\natexlab}[1]{#1}
\providecommand{\url}[1]{\texttt{#1}}
\expandafter\ifx\csname urlstyle\endcsname\relax
  \providecommand{\doi}[1]{doi: #1}\else
  \providecommand{\doi}{doi: \begingroup \urlstyle{rm}\Url}\fi

\bibitem[{B. Zhou et al.}(2018)]{Zhou_2018_ECCV}
{B. Zhou et al.}
\newblock Temporal relational reasoning in videos.
\newblock In \emph{ECCV}, 2018.

\bibitem[Ben-Yosef and Ullman(2018)]{BenYosef2018ImageIA}
Guy Ben-Yosef and Shimon Ullman.
\newblock Image interpretation above and below the object level.
\newblock \emph{Interface focus}, 8 4, 2018.

\bibitem[Ben-Yosef et~al.(2018)Ben-Yosef, Assif, and Ullman]{BENYOSEF201865}
Guy Ben-Yosef, Liav Assif, and Shimon Ullman.
\newblock Full interpretation of minimal images.
\newblock \emph{Cognition}, 171:\penalty0 65 -- 84, 2018.
\newblock ISSN 0010-0277.
\newblock \doi{https://doi.org/10.1016/j.cognition.2017.10.006}.

\bibitem[Ben-Yosef et~al.(2020)Ben-Yosef, Kreiman, and
  Ullman]{BENYOSEF2020104263}
Guy Ben-Yosef, Gabriel Kreiman, and Shimon Ullman.
\newblock Minimal videos: Trade-off between spatial and temporal information in
  human and machine vision.
\newblock \emph{Cognition}, 201:\penalty0 104263, 2020.
\newblock ISSN 0010-0277.
\newblock \doi{https://doi.org/10.1016/j.cognition.2020.104263}.

\bibitem[Breiman(2001)]{Breiman}
L.~Breiman.
\newblock Machine learning.
\newblock In \emph{Random forests}, 2001.

\bibitem[et~al.(2018)]{8578773}
D.~{Tran} et~al.
\newblock A closer look at spatiotemporal convolutions for action recognition.
\newblock In \emph{CVPR}, 2018.
\newblock \doi{10.1109/CVPR.2018.00675}.

\bibitem[et~al.(2020)]{Stroud_2020_WACV}
J.~Stroud et~al.
\newblock D3d: Distilled 3d networks for video action recognition.
\newblock In \emph{IEEE WACV}, 2020.

\bibitem[et~al(2020)]{9062552}
J.~{Zhang} et~al.
\newblock Temporal reasoning graph for activity recog.
\newblock \emph{IEEE Tran. Image Proc.}, 29:\penalty0 5491--5506, 2020.
\newblock \doi{10.1109/TIP.2020.2985219}.

\bibitem[Goyal et~al.(2017)Goyal, Ebrahimi~Kahou, Michalski, Materzynska,
  Westphal, Kim, Haenel, Fruend, Yianilos, Mueller-Freitag,
  et~al.]{goyal2017something}
Raghav Goyal, Samira Ebrahimi~Kahou, Vincent Michalski, Joanna Materzynska,
  Susanne Westphal, Heuna Kim, Valentin Haenel, Ingo Fruend, Peter Yianilos,
  Moritz Mueller-Freitag, et~al.
\newblock The "something something" video database for learning and evaluating
  visual common sense.
\newblock In \emph{Proceedings of the IEEE International Conference on Computer
  Vision}, pages 5842--5850, 2017.

\bibitem[{Jiang} et~al.(2019){Jiang}, {Wang}, {Gan}, {Wu}, and {Yan}]{9010925}
B.~{Jiang}, M.~{Wang}, W.~{Gan}, W.~{Wu}, and J.~{Yan}.
\newblock Stm: Spatiotemporal and motion encoding for action recognition.
\newblock In \emph{2019 IEEE/CVF International Conference on Computer Vision
  (ICCV)}, pages 2000--2009, 2019.
\newblock \doi{10.1109/ICCV.2019.00209}.

\bibitem[{Lin} et~al.(2019){Lin}, {Gan}, and {Han}]{9008827}
J.~{Lin}, C.~{Gan}, and S.~{Han}.
\newblock Tsm: Temporal shift module for efficient video understanding.
\newblock In \emph{2019 IEEE/CVF International Conference on Computer Vision
  (ICCV)}, pages 7082--7092, 2019.
\newblock \doi{10.1109/ICCV.2019.00718}.

\bibitem[Materzynska et~al.(2020)Materzynska, Xiao, Herzig, Xu, Wang, and
  Darrell]{CVPR2020_SomethingElse}
Joanna Materzynska, Tete Xiao, Roei Herzig, Huijuan Xu, Xiaolong Wang, and
  Trevor Darrell.
\newblock Something-else: Compositional action recognition with
  spatial-temporal interaction networks.
\newblock In \emph{CVPR}, 2020.

\bibitem[Wang and Gupta(2018)]{wang2018videos}
Xiaolong Wang and Abhinav Gupta.
\newblock Videos as space-time region graphs.
\newblock In \emph{Proceedings of the European conference on computer vision
  (ECCV)}, pages 399--417, 2018.

\bibitem[Wu et~al.(2019)Wu, Feichtenhofer, Fan, He, Krahenbuhl, and
  Girshick]{Wu_2019_CVPR}
Chao-Yuan Wu, Christoph Feichtenhofer, Haoqi Fan, Kaiming He, Philipp
  Krahenbuhl, and Ross Girshick.
\newblock Long-term feature banks for detailed video understanding.
\newblock In \emph{Proceedings of the IEEE/CVF Conference on Computer Vision
  and Pattern Recognition (CVPR)}, June 2019.

\end{thebibliography}
\end{document}